\documentclass[10pt,twocolumn,letterpaper]{article}

\usepackage[review]{cvpr}      
\usepackage[dvipsnames]{xcolor}

\usepackage{times}
\usepackage{epsfig}
\usepackage{graphicx}
\usepackage{amsmath}
\usepackage{amssymb}

\usepackage{multirow}
\usepackage{tabularx}
\usepackage{booktabs}
\usepackage{longtable}

\usepackage[accsupp]{axessibility}  
\usepackage[font=footnotesize,labelfont=bf]{caption}
\usepackage[belowskip=-5pt,aboveskip=3pt]{caption} 

\usepackage{dsfont}
\usepackage{pifont}
\usepackage{url}
\usepackage{color}
\usepackage{amsthm}
\usepackage[export]{adjustbox}
\usepackage{comment}
\usepackage{algorithm} 

\usepackage[utf8]{inputenc} 
\usepackage[T1]{fontenc}    
\usepackage{url}            
\usepackage{amsfonts}       
\usepackage{nicefrac}       
\usepackage{microtype}      
\usepackage{enumitem}
\usepackage{blindtext}
\usepackage{sidecap}
\usepackage{wrapfig}

\usepackage{dsfont}

\definecolor{dg}{rgb}{0,0.694,0.298}
\definecolor{purple}{rgb}{0.4,0.176,0.569}
\definecolor{royalblue}{RGB}{65,105,225}
\usepackage{pifont}
\newcommand{\figref}[1]{Fig.~\ref{#1}}
\newcommand{\reqref}[1]{Eq.~(\ref{#1})}
\newcommand{\secref}[1]{Sec.~\ref{#1}}

\makeatletter
\DeclareRobustCommand\onedot{\futurelet\@let@token\@onedot}
\def\@onedot{\ifx\@let@token.\else.\null\fi\xspace}
\def\eg{\emph{e.g}\onedot} 
\def\ie{\emph{i.e}\onedot}

\makeatother

\definecolor{americanrose}{rgb}{1.0, 0.01, 0.24}

\usepackage{algpseudocode} 
\usepackage{tikz}

\definecolor{cvprblue}{rgb}{0.21,0.49,0.74}
\usepackage[pagebackref,breaklinks,colorlinks,citecolor=cvprblue]{hyperref}

\title{\textbf{\textsc{C-NeRF}: Representing Scene Changes as Directional Consistency Difference-based NeRF}}

\author{Rui Huang$^{1}$ \quad Binbin Jiang$^{1}$ \quad Qingyi Zhao$^{1}$ \quad William Wang \quad Yuxiang Zhang$^{1}$ \quad Qing Guo$^{2,3, *}$ \\
$^{1}$ College of Computer Science and Technology, Civil Aviation University of China, China\\
$^{2}$ IHPC, Agency for Science, Technology and Research, Singapore\\
$^{3}$ CFAR, Agency for Science, Technology and Research, Singapore
}

\begin{document}



\twocolumn[{
\renewcommand\twocolumn[1][]{#1}
\date{}
\maketitle
\begin{center}
    \captionsetup{type=figure}
    \includegraphics[width=\textwidth]{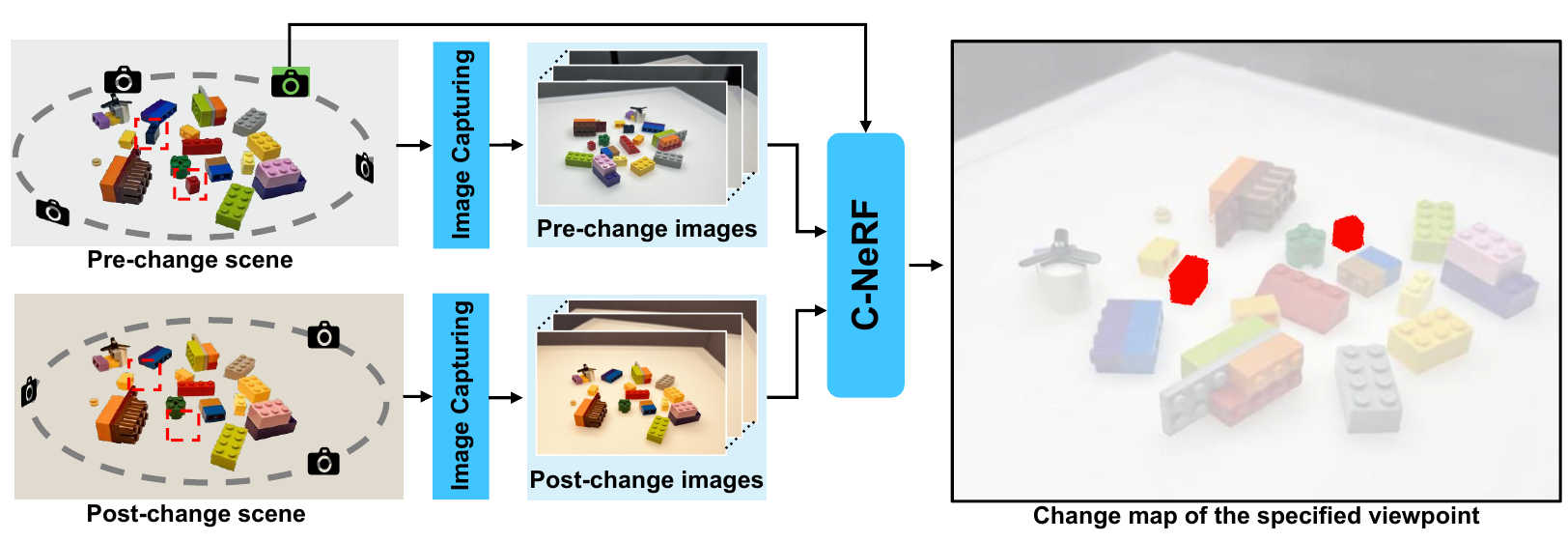}
    \captionof{figure}{Overview of \textsc{C-NeRF}. We aim to build a representation for 3D changes of a scene from multiview images captured before and after object changes. \textsc{C-NeRF} can generate change maps based on arbitrarily specified views (see the left image). We highlight the changes in the scene via dashed red rectangles.} \vspace{10px}
    \label{fig:overview}
\end{center}
}]

\begin{abstract}
In this work, we aim to detect the changes caused by object variations in a scene represented by the neural radiance fields (NeRFs). 
Given an arbitrary view and two sets of scene images captured at different timestamps, we can predict the scene changes in that view, which has significant potential applications in scene monitoring and measuring. 
We conducted preliminary studies and found that such an exciting task cannot be easily achieved by utilizing existing NeRFs and 2D change detection methods with many false or missing detections. 
The main reason is that the 2D change detection is based on the pixel appearance difference between spatial-aligned image pairs and neglects the stereo information in the NeRF.
To address the limitations, we propose the \textsc{C-NeRF} to represent scene changes as directional consistency difference-based NeRF, which mainly contains three modules.
We first perform the spatial alignment of two NeRFs captured before and after changes. 
Then, we identify the change points based on the direction-consistent constraint; that is, real change points have similar change representations across view directions, but fake change points do not.
Finally, we design the change map rendering process based on the built NeRFs and can generate the change map of an arbitrarily specified view direction.
To validate the effectiveness, we build a new dataset containing ten scenes covering diverse scenarios with different changing objects. 
Our approach surpasses state-of-the-art 2D change detection and NeRF-based methods by a significant margin. Our code is available at \href{ https://github.com/C-NeRF/C-NeRF} {https://github.com/C-NeRF/C-NeRF}
%
\end{abstract}

\section{Introduction}
Change detection aims to identify regions or objects that have changed in the scene, which has various applications in fields such as remote sensing \cite{chen2020spatial,khan2017forest}, surveillance \cite{huwer2000adaptive,zhang2020extended}, and robot navigation \cite{takeda2023lifelong, krawciw2023change}. 
Previous works mainly detect scene changes based on spatially aligned image pairs that are captured by a camera at two timestamps.
Although such 2D methods achieve significant progress \cite{huang2023background}, they have two inherent limitations: 
\textit{First}, 2D change detections could only identify changes in one view direction. Some changes hidden in other objects caused by occlusion could not be identified. 
\textit{Second}, the 3D shape of the changing objects cannot be known, which limits the potential applications significantly. A 3D change detector should be developed.

%
\begin{figure*}[tb]
  \centering
\includegraphics[width=\linewidth]{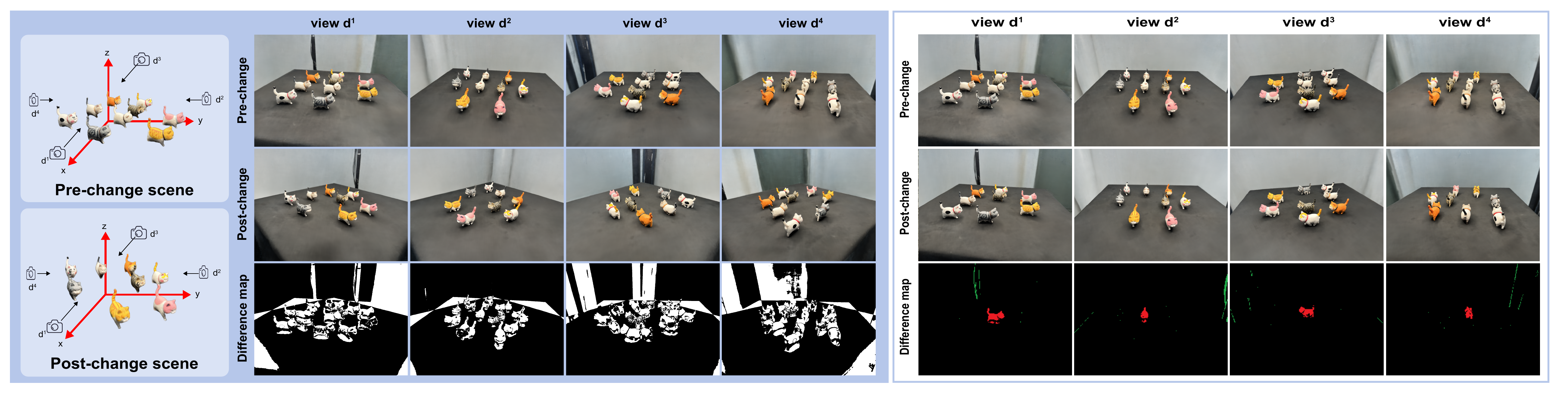}
  \caption{Left: images generated from two independent NeRF models with identical spatial location and viewing direction. Right: images generated from two spatially aligned NeRFs. The {\color[RGB]{251,55,49}{red}} points represent real changes. The {\color[RGB]{19,138,7}{green}} points indicate fake changes.}\vspace{-10px}
  \label{fig:motivation}
\end{figure*}

In this work, we propose to extend the 2D change detection task to 3D based on the neural radiance fields (NeRFs) \cite{mildenhall2020nerf,chen2022aug}. 
Compared with other 3D representations (\eg, point-clouds \cite{aliev2020neural}, meshes \cite{riegler2020free,thies2019deferred}, voxels \cite{lombardi2019neural}), NeRF is able to represent a 3D scene from multi-view images and synthesize photo-realistic novel views by learning a multi-layer perceptron (MLP) to map an input 5D coordinates (\ie, spatial location and 2D view direction) to the volumen density and view-dependent emitted radiance color, which has be extended to diverse challenging scenarios \cite{tancik2022block,pumarola2021d}. 

With the NeRF \cite{mildenhall2020nerf} for 3D representation, the change detection in 3D can be formulated as a change map synthesis task: given an arbitrary view and two sets of scene images captured at different timestamps, we aim to predict a binary change map indicating the changing regions in that view, as shown in Fig.~\ref{fig:overview}.
However, such a task is non-trivial and cannot be simply achieved through existing NeRF methods.
In particular, a naive solution is to build two independent NeRFs based on the two image sets and synthesize two view images by feeding the two NeRFs with the same view direction.
Then, we get two synthesized images and can calculate the difference between two images for change detection.
As shown in \figref{fig:motivation}, such a method easily leads to a lot of false or missing detections caused by: 
\ding{182} The two NeRFs of the 3D scene before and after change are not registered in the same coordinate. This means that even if the spatial location and viewing direction are identical for the NeRF models, the generated images are still not aligned, which results in a lot of false detections (see \figref{fig:motivation} (Left)). 
\ding{183} Even if we have two aligned images, the absolute difference of the image pair is also affected by rendering quality and environment differences (\eg, light changes in \figref{fig:overview} and \figref{fig:motivation} (Right)).


To address the limitations, we propose  \textsc{C-NeRF} to represent scene changes as directional consistency difference-based NeRF, which mainly contains three modules. 
We first perform the spatial alignment of two NeRFs captured before and after changes. 
Then, we identify the change points based on the direction-consistent constraint; that is, real change points have similar change representations across view directions, but fake change points do not. 
Finally, we design the change map rendering process based on the built NeRFs and can generate the change map of an arbitrarily specified view direction. 
To validate the effectiveness, we build a new dataset containing ten scenes covering diverse scenarios with different changing objects. 
In summary, the main contributions of our paper are as follows:
\begin{itemize}
    \item We extend the 2D change detection task to 3D, and propose \textsc{C-NeRF} by predicting the scene changes of an arbitrary view from two sets of scene images captured at different timestamps. 
    With such a novel method, we can synthesize change maps under arbitrary specified view directions, which indicates the change regions under those views. 

    \item We observe that the real change points have similar change representations across different view directions but the fake change points do not.
    Inspired by this, we propose a real change points identification method based on the direction-consistent constraint, which is able to identify instance-level and fine-grained changes effectively.
    
    \item To validate the effectiveness, we build a new scene change detection dataset containing ten scenes covering diverse scenarios with different change objects and compare with a series of baselines.
    
\end{itemize}

\section{Related Work}
\textbf{2D change detection methods.} 
2D change detection has been widely studied in image change captioning, remote sensing applications and street view scenes \cite{2d_captioning_jhamtani2018learning, chen2020spatial, huang2023background, sakurada2013detecting, alcantarilla2018street}. 
%
%
The basic requirement of these methods is that images of before and after scene change have limited viewpoint shifts.

Image change captioning aims to locate and describe changed objects by using natural language \cite{2d_captioning_oluwasanmi2019fully}. Huang et al. \cite{2d_captioning_huang2021image} propose a method to accurately locate changed objects using fine-grained features such as visual, semantic, and positional features at the instance-level. 
Kim et al. \cite{2d_captioning_kim2021agnostic} present a viewpoint-agnostic change captioning network with cycle consistency to improve the robustness of the change detection captured with large camera perspectives. 
Qiu et al. \cite{2d_captioning_qiu2021describing} propose multi-change captioning transformers that identify change regions by densely correlating different regions in image pairs and dynamically determining the related change regions with words in sentences. 
Tu et al. \cite{2d_captioning_tu2021semantic} propose a semantic relation-aware difference representation learning network that explicitly learns the difference representation in the presence of illumination or viewpoint change.
Unlike change captioning, change detection in remote sensing or street view scenes always outputs binary maps indicating the changed pixel. 
In \cite{jst2015change}, Sakurada et al. use a convolutional neural network in combination with superpixel segmentation to detect changes on a pair of street-view images. 
Lei et al. \cite{lei2020hierarchical}  design an effective feature fusion method to improve the accuracy of the corresponding change maps. 
Chen et al. \cite{chen2021dr} propose the temporal attention and explore the impact of the dependency-scope size of temporal attention on the performance of change detection. 
Sachdeva et al. \cite{sachdeva2023change2D} aims to detect the``object-level'' change in an image pair with different illumination at different viewpoints. 
In their latest work \cite{sachdeva2023change3D}, the authors' further study change detection with a significant shift in camera pose. 
2D change detection requires {image alignment}, training with a large volume of change image pairs, and can only generate change maps at { the viewpoint of the input images}. 
Obtaining change maps at different viewpoints can be challenging.

\textbf{3D change detection.}
3D change detection can reflect change in 3D space with multi-view images of before and after scene change and 3D scans \cite{taneja2011image,ulusoy2014image,6942806,qin20163d}. 
Previous research \cite{5980542} has utilized dense color and depth information to detect changes between 3D maps. 
Fehr et al. \cite{7989614} develop a 3D reconstruction algorithm based on an extended Truncated Signed Distance Function to more accurately solve the scene differencing problem. 
Ku et al. \cite{ku2021shrec} have contributed a street-scene dataset for 3D point cloud change detection that can detect changes in a complex street environment from multi-temporal point clouds. 
Qiu et al. \cite{qiu20233d} propose a method to explicitly localize changes in 3D bounding boxes from two point clouds, describing detailed scene changes. 
In contrast with existing methods, our method only requires multi-view images before and after scene change, {without camera pose alignment}. 
In addition, our method is capable of {capturing the change in 3D space and rendering change maps at any given viewpoint}.
%

\textbf{NeRF for scene representation.} 
Neural Radiance Field (NeRF) \cite{mildenhall2020nerf} is a 3D scene representation technology that uses a deep neural network to generate novel views of a 3D scene based on the corresponding view directions. 
Several methods have been proposed to improve the quality and speed of NeRF models. 
Barron et al., \cite{barron2021mip} propose a method to represent the scene at a continuously-valued scale and render anti-aliased conical frustums instead of rays, which improves the fine details of the generated image. 
Niemeyer et al., \cite{niemeyer2022regnerf} regularize the geometry and appearance of patches rendered from unobserved viewpoints to enhance the NeRF. 
Zhang et al., \cite{zhang2022ray} propose a random ray casting policy that enables training unseen views using seen views to produce high-quality renderings under novel viewpoints. 
Slow rendering times always hinder the use of NeRF. 
To achieve high rendering speed, FastNeRF \cite{garbin2021fastnerf} and KiloNeRF \cite{reiser2021kilonerf} propose using multiple MLPs.
Instant-NGP\cite{muller2022instant} propose to use multi-resolution hash encoding as network input to reduce the number of floating point and memory access operations. 
EfficientNeRF\cite{hu2022efficientnerf} proposes using multi-resolution hash encoding as network input to reduce the number of floating-point and memory access operations. 
MobileNeRF\cite{chen2023mobilenerf} introduces a new NeRF representation based on textured polygons that can synthesize novel images efficiently with standard rendering pipelines. 
Different from these works improving the rendering quality and speed of NeFR models, our method is built on NeRF and solves change detection {without camera pose alignment}.

\section{Preliminary: Neural Radiance Fields}
\label{sec:preliminary}

Neural radiance fields (NeRF) \cite{mildenhall2020nerf} can synthesize new views via a MLP (multilayer perception) $\Psi$ trained with a set of sparse images $\mathcal{I} = \{\mathbf{I}_1, \mathbf{I}_2, \dots, \mathbf{I}_N\}$ captured at different views, which can be denoted as 
$\Psi = \Gamma_\text{train}(\mathcal{I}, \mathcal{D})$. The set $\mathcal{D}=\{\mathbf{d}_1, \mathbf{d}_2,\ldots,\mathbf{d}_N\}$ contains view directions of images in $\mathcal{I}$, which are calculated by the method \cite{schonberger2016structure}.
During the inference process, NeRF takes a 5D coordinate including the spatial location $\mathbf{x}=(x, y, z)$ and the viewing direction $\mathbf{d}$ = ($\theta$, $\phi$) as inputs and can estimate the volume density $\sigma$ and view-dependent emitted radiance color $\mathbf{c}=(r,g,b)$ of the specified coordinate, that is $(\mathbf{c},\sigma) = \Psi(\mathbf{x},\mathbf{d})$.
%

When we render an image under the view direction $\mathbf{d}$, for a pixel $\mathbf{p}$, we have a ray that is shot from the camera center through the pixel, and we can estimate its color by
$\mathbf{I}[\mathbf{p}] = \sum_{i=1}^K    T_i\alpha_i\mathbf{c}_i$,
%
%
%
where $\{\mathbf{x}_i\}_{i=1}^{K}$ contains $K$ points from the ray. The variables $\alpha_i$ and $T_i$ are the alpha values for blending and the accumulated transmittance, which are determined by the estimated volume density. We denote the whole render process for the given view direction $\mathbf{d}$ as $\mathbf{I} = \Gamma_\text{render}(\Psi,\mathbf{d})$.
Please refer to \cite{mildenhall2020nerf} for more details.

\begin{figure*}[!htb]
  \centering
  \includegraphics[width=\linewidth]{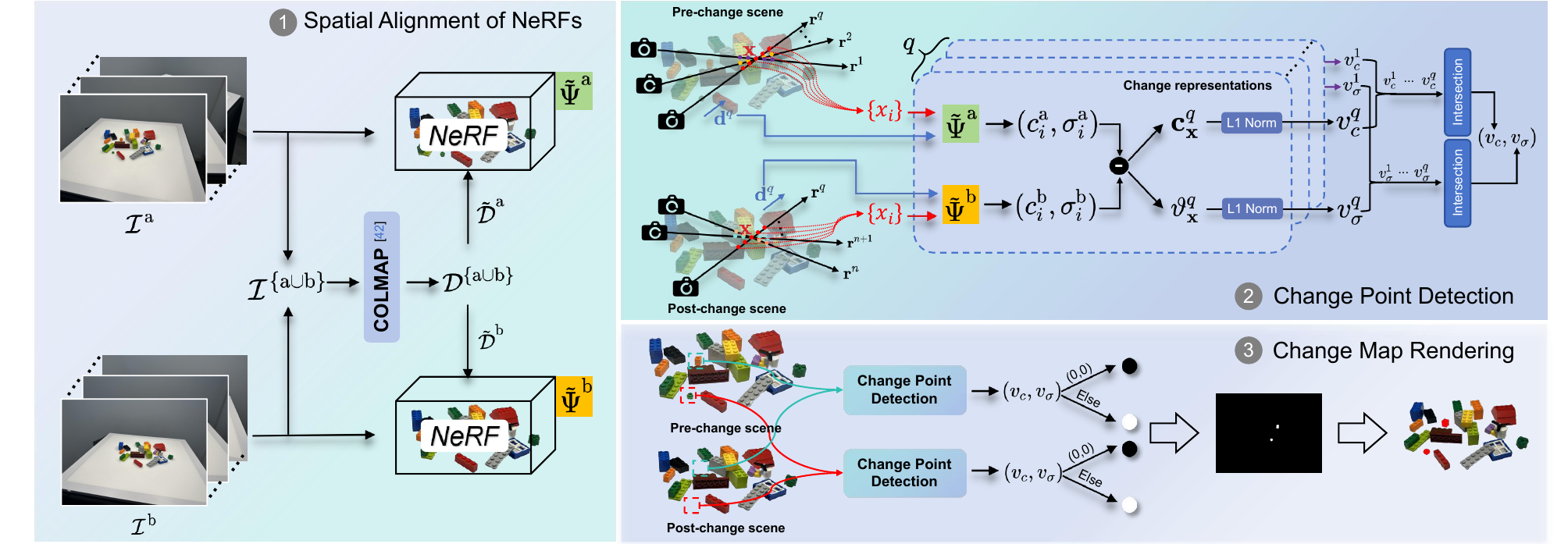}
  \caption{The main framework of the proposed \textsc{C-NeRF}.} \vspace{-10px}
  \label{fig:framework}
\end{figure*}

\section{Scene Change Representation}
\label{sec:problem_mot}

\subsection{Problem Formulation}
\label{subsec:problem}

For a scene we want to monitor, we collect a set of images $\mathcal{I}^\text{a}=\{\mathbf{I}^\text{a}_1,\mathbf{I}^\text{a}_2,\ldots,\mathbf{I}^\text{a}_{N_a}\}$ at the first time observation (\ie, pre-change images). After some time, we observe the scene again and get another set of images $\mathcal{I}^\text{b}=\{\mathbf{I}^\text{b}_1,\mathbf{I}^\text{b}_2,\ldots,\mathbf{I}^\text{b}_{N_b}\}$ (\ie, post-change images).
Our goal is to learn a mapping that can represent the changes in the scene at arbitrary viewing direction, in particular, the fine-grained changes, between two observations. 
Such a requirement is expected in the real world. For example, we would like to monitor and measure the scene changes around high-value objects, such as cultural relics and jewelry, to protect them.
However, existing 2D change detections cannot achieve the goal since they usually require the images captured two times to be spatially aligned.

Specifically, given a 3D point $\mathbf{x} = (x,y,z)$ and a view direction $\mathbf{d}=(\theta,\phi)$, a learned representation can estimate whether the point is a changed point or not
\begin{align}\label{eq:cd}
    (v_c,v_\sigma) = \Psi^v(\mathbf{x},\mathbf{d}),
\end{align}
where $v_c \in \{0,1\} $ and $v_\sigma \in \{0,1\}$ are two change indicators indicating whether the color and volume density of $\mathbf{x}$ change or not dependent on $\mathbf{d}$.
Different from learning $\Psi$ within \secref{sec:preliminary}, $\Psi^{v}$ should depend on the two observations with two sparse sets, \ie, $\mathcal{I}^\text{a}$ and $\mathcal{I}^\text{b}$.


\subsection{Naive Solution}
\label{subsec:solution}

We can employ existing NeRF methods and calculate appearance and density difference to realize the task defined in \reqref{eq:cd}. Specifically, we first use an existing NeRF method to build NeRF models based on the two sets $\mathcal{I}^\text{a}$ and $\mathcal{I}^\text{b}$, respectively, \ie,
\begin{align}\label{eq:cd_naive_1}
    \Psi^\text{a} = \Gamma_\text{train}(\mathcal{I}^\text{a},\mathcal{D}^\text{a}),~\text{and} ~\Psi^\text{b} = \Gamma_\text{train}(\mathcal{I}^\text{b},\mathcal{D}^\text{b}), 
\end{align}
where $\mathcal{D}^\text{a}$ and $\mathcal{D}^\text{b}$ contain directions of images in $\mathcal{I}^\text{a}$ and $\mathcal{I}^\text{b}$, respectively. We calculate $\mathcal{D}^\text{a}$ and $\mathcal{D}^\text{b}$ by feeding $\mathcal{I}^\text{a}$ and $\mathcal{I}^\text{b}$ to \cite{schonberger2016structure}, respectively.
Then, we estimate the color of the same point under a view direction via the two NeRF models, respectively, and get
\begin{align}\label{eq:cd_naive_2}
    (c^\text{a},\sigma^\text{a}) = \Psi^\text{a}(\mathbf{x},\mathbf{d}),  
    ~(c^\text{b},\sigma^\text{b}) = \Psi^\text{b}(\mathbf{x},\mathbf{d}).
\end{align}

After that, we can calculate the difference between the two colors and get
\begin{align}\label{eq:cd_naive_3}
   c^\text{a-b} = |c^\text{a} - c^\text{b}|, ~\sigma^\text{a-b} = |\sigma^\text{a} - \sigma^\text{b}|.
\end{align}

Intuitively, we can calculate $v$ via a threshold for
$c^\text{a-b}$,
\begin{align}\label{eq:cd_naive_4}
   v_c = \left\{\begin{matrix} 
        1, c^\text{a-b}>\epsilon_c \hfill\hfill\\
        0, \text{otherwise}
   \end{matrix}\right.,
    ~ v_\sigma = \left\{\begin{matrix} 
        1, \sigma^\text{a-b}>\epsilon_\sigma\\
        0, \text{otherwise}
   \end{matrix}\right..
\end{align}

Although easy, this strategy requires the two built NeRF models to be well-aligned in both spatial space and appearance. That is, given the same direction, the $\Psi^\text{a}$ and $\Psi^\text{b}$ should render the same view and have the same color at the same point. 
Nevertheless, $\Psi^\text{a}$ and $\Psi^\text{b}$ are built independently and usually cannot generate the same view with a given direction. We show an example in \figref{fig:motivation} (Left) where we build two NeRFs for a scene respectively and render a view by specifying the same direction. 
The rendered views are different and lead to numerous fake changes.
In addition, the difference caused by environmental changes (\eg, light changes) and rendering quality instead of object changes will also lead to a high $c^\text{a-b}$ and affect the change detection results significantly. As the case is shown in \figref{fig:motivation} (Right), even though the two NeRF models are well aligned, there are a lot of fake changes (\ie, green points), and real changes are missed.

\section{Methodology: \textsc{C-NeRF}}
\label{sec:method}

Instead of detecting changes based on differences between two rendered images, we argue that a real change point should have distinct color differences in all view directions.
To this end, we propose aligning two NeRF models in the same coordinate system through structure-from-motion (SFM), as detailed in \secref{subsec:sfm-alignment}. Then, we propose the \textit{direction-consistent change point detection} in \secref{subsec:change-identification} to filter fake changes and identify fine-grained changes automatically. Finally, we introduce how to render the change map with the specified direction in \secref{subsec:change-rendering}. The main framework of \textsc{C-NeRF} is shown in Fig.~\ref{fig:framework}.

\subsection{Spatial Alignment of NeRFs}
\label{subsec:sfm-alignment}

Given two image sets captured at different timestamps (\ie, $\mathcal{I}^\text{a}$ and $\mathcal{I}^\text{b}$), we merge them and get a set $\mathcal{I}^{\{\text{a}\cup\text{b}\}}=\{\mathcal{I}^\text{a},\mathcal{I}^\text{b}\}$. 
Different from the \reqref{eq:cd_naive_1} where the two NeRF models are trained with independently calculated $\mathcal{D}^\text{a}$ and $\mathcal{D}^\text{b}$, we use the method \cite{schonberger2016structure} to assign directions for all images in $\mathcal{I}^{\{\text{a}\cup\text{b}\}}$ as a whole and get $\mathcal{D}^{\{\text{a}\cup \text{b}\}}=\{\tilde{\mathcal{D}}^\text{a},\tilde{\mathcal{D}}^\text{b}\}$.
As a result, the two image sets are aligned in the same coordinate system.
With the new direction sets $\{\tilde{\mathcal{D}}^\text{a},\tilde{\mathcal{D}}^\text{b}\}$, we can use \reqref{eq:cd_naive_1} to rebuild the two NeRF models denoted as $\tilde{\Psi}^\text{a}$ and $\tilde{\Psi}^\text{b}$. 

With such a simple strategy, the two NeRF models captured at two different timestamps are well-aligned in the spatial space, which eliminates the fake changes significantly.
However, the fake changes caused by environmental changes and rendering quality cannot be appropriately addressed.

\subsection{Direction-Consistent Change Point Detection}
\label{subsec:change-identification}

A real change point $\mathbf{x}$ should have the following properties: 
\ding{182} Its appearances or densities estimated from the two NeRF models are usually distinct. 
%
\ding{183} The first property should be consistent across different view directions.
To achieve the two properties, we design a change point detection method based on the built $\tilde{\Psi}^\text{a}$ and $\tilde{\Psi}^\text{b}$.

\begin{figure}[!tb]
  \centering
  \includegraphics[width=\linewidth]{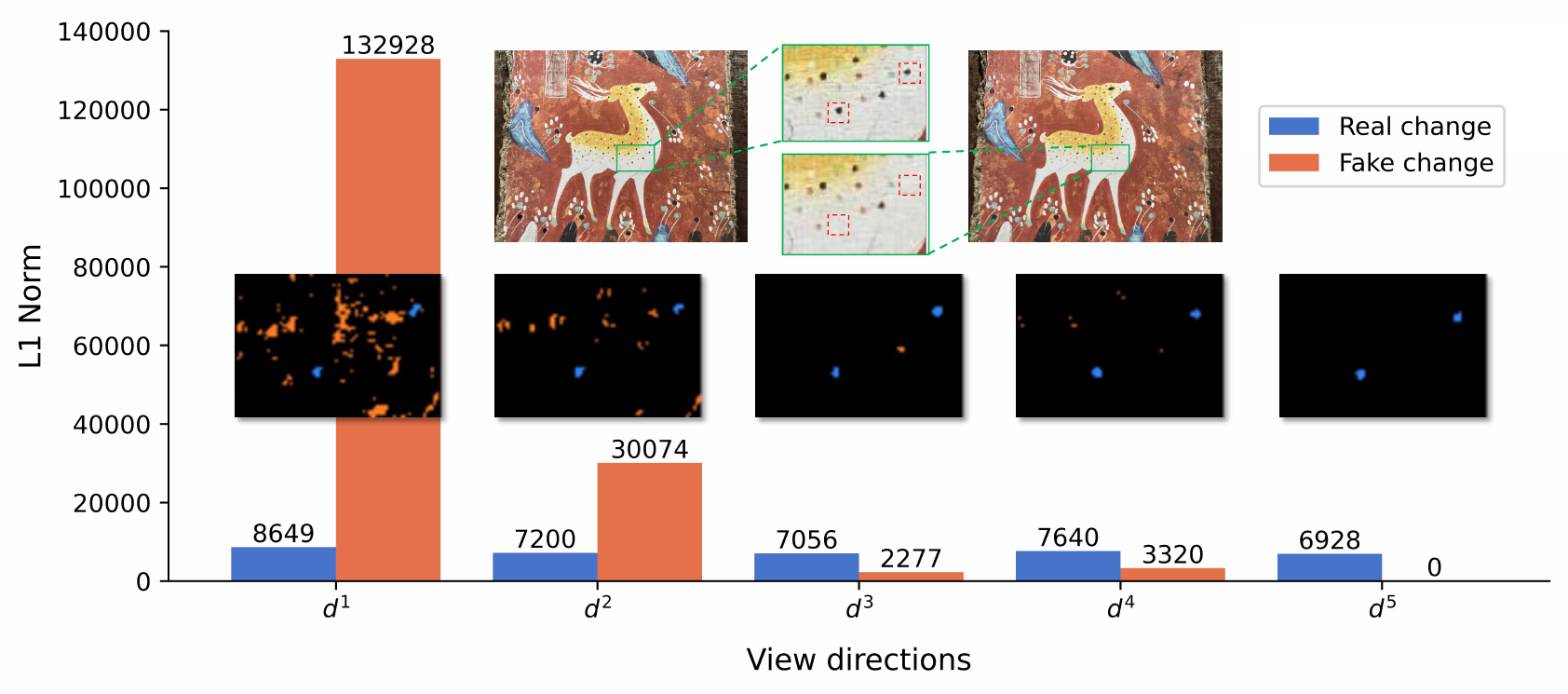}
  \caption{Total
L1 norms of real and fake change points of the green box on five view directions.}\vspace{-10px}
  \label{fig:statics}
\end{figure}

\textbf{Change representations.} 
Given a point $\mathbf{x}$ and $q$th view direction $\mathbf{d}^q$ that can see the point, we have a ray $\mathbf{r}^q$ that is shot from the camera center to the point $\mathbf{x}$. 
We sample points near $\mathbf{x}$ along the ray $\mathbf{r}^q$ and get $\mathcal{N}=\{\mathbf{x}_i\}$.
Then, we calculate the color difference of $\mathbf{x}_i\in\mathcal{N}$ based on $\tilde{\Psi}^\text{a}$ and $\tilde{\Psi}^\text{b}$, and use their weighted difference as the representation of the appearance change of the point $\mathbf{x}$ under the view $\mathbf{d}^q$,
\vspace{-0.38cm}
\begin{align} \label{eq:diff}
     & \mathbf{c}_{\mathbf{x}}^q = [c^{\text{a}-\text{b}}_1,c^{\text{a}-\text{b}}_2,\ldots,c^{\text{a}-\text{b}}_{|\mathcal{N}|}],\\
     & \vartheta_{\mathbf{x}}^q = [\sigma^{\text{a}-\text{b}}_1,\sigma^{\text{a}-\text{b}}_2,\ldots,\sigma^{\text{a}-\text{b}}_{|\mathcal{N}|}],\\
    ~~\text{s.t.}~~ & c^{\text{a}-\text{b}}_i = |T_i^\text{a}\alpha_i^\text{a}c_i^\text{a} - T_i^\text{b}\alpha_i^\text{b} c_i^\text{b}|, \nonumber \\
    & \sigma^{\text{a}-\text{b}}_i  = |T_i^\text{a}\alpha_i^\text{a}\sigma_i^a - T_i^\text{b}\alpha_i^\text{b}\sigma_i^b|, \nonumber
\end{align}
where $(c_i^\text{a}, \sigma_i^\text{a} )= \tilde{\Psi}^\text{a}(\mathbf{x}_i,\mathbf{d}^q)$, and $(c_i^\text{b}, \sigma_i^\text{b}) = \tilde{\Psi}^\text{b}(\mathbf{x}_i,\mathbf{d}^q)$ with $i \in [1,\ldots,|\mathcal{N}|]$.
In addition, we follow the sampling strategy used in the original NeRF and have $T_i^* = \mathrm{exp} (-\sum_{j=1}^{i-1}\sigma_j^\text{*} \delta_j)$,
$\alpha_i^* = 1-\mathrm{exp} (\sigma_i^* \delta_i), *\in\{a,b\}$, and $\delta_i$ denotes the distance between two neighboring sampled points.
Intuitively, a real change point $\mathbf{x}$ easily leads to distinct pre-post changes of neighboring points around the $\mathbf{x}$ itself, and we consider the influence of point density by involving $\{T_i^\text{*}\alpha_i^\text{*}|*\in\{a,b\}\}$ in \reqref{eq:diff}.
Then, we use the weighted pre-post differences of neighboring points to represent appearance changes of $\mathbf{x}$ and calculate the change indicators of $\mathbf{x}$ in \reqref{eq:cd} based on the $q$th view by
\begin{align}\label{eq:cnerf-1}
    v^q_c = \left\{\begin{matrix} 
        1, \|\mathbf{c}_\mathbf{x}^q\|_1>\epsilon_c \\
        0, \text{otherwise} \hfill\hfill
   \end{matrix}\right.,
    ~ v^q_\sigma = \left\{\begin{matrix} 
        1, \|\vartheta_\mathbf{x}^q\|_1>\epsilon_\sigma \\
        0, \text{otherwise} \hfill\hfill
   \end{matrix}\right..
\end{align}

However, based on a single view, the representation is easily disturbed by rendering noise or environmental changes. As an example shown in \figref{fig:motivation} (Right), there are numerous fake changes based on the \reqref{eq:cnerf-1}.

\textbf{Direction-consistent constrain.}
We observe that the real change points' representations always have high $L_1$ norms and get consistent $v_c$ or $v_\sigma$ across different views. 
In contrast, the fake change points' representations only have high $L_1$ norms in a few views and low $L_1$ norms across other views.
An example is shown in \figref{fig:statics} where we calculate the total $L_1$ norms of real and fake change points of the green box on five view directions. 
Clearly, the real change points have similar $L_1$  norms across different directions while the fake ones have significantly large $L_1$ norm on $\mathbf{d}_1$ and zero $L_1$ norm on $\mathbf{d}_5$.
Specifically, given a set of view directions $\tilde{\mathcal{D}}=\{\mathbf{d}^q\}$, we can compute the representation of $\mathbf{x}$ under each view direction and get $\{(v^q_c, v^q_\sigma)| \mathbf{d}^q\in\tilde{\mathcal{D}}\}$ via \reqref{eq:cnerf-1}.
According to the direction consistent fact, if $\mathbf{x}$ is a real change point caused by color variation (or density variation), all $\{v^q_c\}$ (or $\{v^q_\sigma\}$) should be one and we can calculate the two indicators of $\mathbf{x}$ as
\vspace{-0.20cm}
\begin{align}\label{eq:cnerf-2}
    (v_c, v_\sigma)= \Psi^v(\mathbf{x},\mathbf{d})= (\bigcap_{q=1}^{|\tilde{\mathcal{D}}|} v^q_c, \bigcap_{q=1}^{|\tilde{\mathcal{D}}|} v^q_\sigma).
\end{align}

Then, the key problem becomes how to collect view directions to build $\tilde{\mathcal{D}}$ to calculate $v$.

\begin{figure}[!tb]
     \centering
     \begin{subfigure}[b]{0.45\linewidth}
         \centering
         \includegraphics[width=\textwidth]{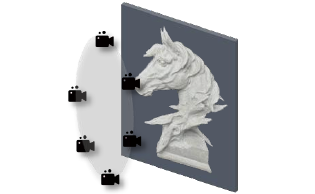}
         \caption{forward-facing scene}\vspace{5px}
         \label{fig:forward-facing render}
     \end{subfigure}
     \hfill
     \begin{subfigure}[b]{0.45\linewidth}
         \centering
         \includegraphics[width=\textwidth]{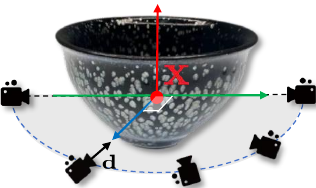}
         \caption{surrounding scene}\vspace{5px}
         \label{fig:surround render}
     \end{subfigure}
        \caption{Examples of $\tilde{\mathcal{D}}$ for forward-facing and surrounding scenes.}\vspace{-10px}
        \label{fig:dataset_construction}
\end{figure}

\textbf{View direction sampling strategies.} 
We follow two principles to select view directions: \ding{182} the desired point $\mathbf{x}$ could be seen along the selected views. \ding{183} As many views as possible should be included.
Moreover, we need to consider two scenarios: 
\textit{Forward-facing scene}, the camera always faces the scene, and each point can be seen in arbitrary view directions. Hence, we can uniformly sample $V$ view directions to form $\tilde{\mathcal{D}}$.
In \figref{fig:forward-facing render}, we show the sampled view directions to calculate $v$.
\textit{Surround scene}, for a point $\mathbf{x}$, the view directions that cannot see the point due to the self-occlusion should be excluded from the set $\tilde{\mathcal{D}}$. 
We take the view direction $\mathbf{d}$ as the center line and rotate the direction from -90 degree to 90 degree centered on the point $\mathbf{x}$ as the sampling range. This range is able to cover most of the view directions that can see the point $\mathbf{x}$.
Then, within this range, we uniformly sample $V$ view directions to form the set $\tilde{\mathcal{D}}$. We show an example in \figref{fig:surround render}.

\subsection{Change Map Rendering under New view}
\label{subsec:change-rendering}

Given a view direction $\mathbf{d}$, we aim to render a binary change map $\mathbf{C}$ that indicates the changed pixels under the view direction. 
Specifically, for the pixel $\mathbf{p}$ of the image captured under view $\mathbf{d}$, we have a ray $\mathbf{r}$ shooting from the camera center through the pixel center.
Then, we sample points along the ray as done in NeRF and find the max values $T_i^\text{a}\alpha_i^\text{a}$ and $T_j^\text{b}\alpha_j^\text{b}$ from $\{T_i^\text{a}\alpha_i^\text{a}\}|_{i=1}^{K}$ and $\{T_j^\text{b}\alpha_j^\text{b}\}|_{j=1}^{K}$, respectively.
%

%

We select the point closest to the camera from $T_i^\text{a}\alpha_i^\text{a}$ and $T_j^\text{b}\alpha_j^\text{b}$ as the center point denoted as $\mathbf{x}$ and can determine the binary value of pixel $\mathbf{p}$ by
\begin{align} \label{eq:render}
    \mathbf{C}[\mathbf{p}] = \text{max}(v_c, v_\sigma),~\text{s.t.},~ (v_c, v_\sigma)= \Psi^v(\mathbf{x},\mathbf{d}).
\end{align}
%

\subsection{Implementation Details}

Given two image sets observed at two timestamps (\ie, $\mathcal{I}^\text{a}$ and $\mathcal{I}^\text{b}$) and a view direction $\mathbf{d}$, we adopt the following steps to generate the change map under $\mathbf{d}$: 
\ding{182} We use the method in \secref{subsec:sfm-alignment} to generate two spatially aligned NeRFs, \ie, $\hat{\Psi}^\text{a}$ and $\hat{\Psi}^\text{b}$. 
\ding{183} We build the view direction set $\tilde{\mathcal{D}}$ according to different scenarios as detailed in `view direction sampling strategies.' 
\ding{184} We calculate the change map for each pixel via \reqref{eq:render} in \secref{subsec:change-rendering} and \reqref{eq:cnerf-2} in \secref{subsec:change-identification}.


\textbf{Configuration.} In this paper, we adopt the NeRF model \cite{mildenhall2020nerf} with default parameters. The image size is set to $1008\times756$. All the experiments are conducted on Nvidia RTX 3090Ti. We set $\epsilon_c \in[60,180]$ and $\epsilon_\sigma \in[100,600]$.
\section{Experimental Results}
\label{sec:experiment}

\subsection{Setups}
\label{subsec:setups}

%

%

\textbf{Dataset.}
Most existing change detection datasets are built from coupled images, which cannot support NeRF model training. To validate our method, we build a new dataset \textbf{NeRFCD}. It consists of 2 \textit{surrounding} scenes (\eg, Cats and Block) and 8 \textit{forward-facing} scenes (\eg, Go pieces, Mural, Card, Text, Potting, Desk, sculpture and Desk).

As shown in Fig.~\ref{fig:dataset_construction}, we adopt a ``Zig-zag'' path to shoot about 20 images for the forward-facing scene. To capture the surrounding scene, we use three horizontal circles around the objects: upper, middle and lower. We sample 40-60 shooting positions on each circle. Note that we keep the height of the camera when shooting at the middle circle. The camera always faces the center of the scene for shooting. The motion trajectories of the two shots before and after the change should be as similar as possible.

To train the fine-grained change detectors, we render 120 image pairs for each scene and use our \textsc{C-NeRF} to generate the corresponding change maps. 
Afterward, we select 10 image pairs from the captured image pairs from each scene to build the testing dataset. 
The training set consists of $1, 200$ image pairs, and the testing set consists of 100 image pairs.

\begin{figure}[!tb]
     \centering
     \begin{subfigure}[b]{0.45\linewidth}
         \centering
         \includegraphics[width=\textwidth]{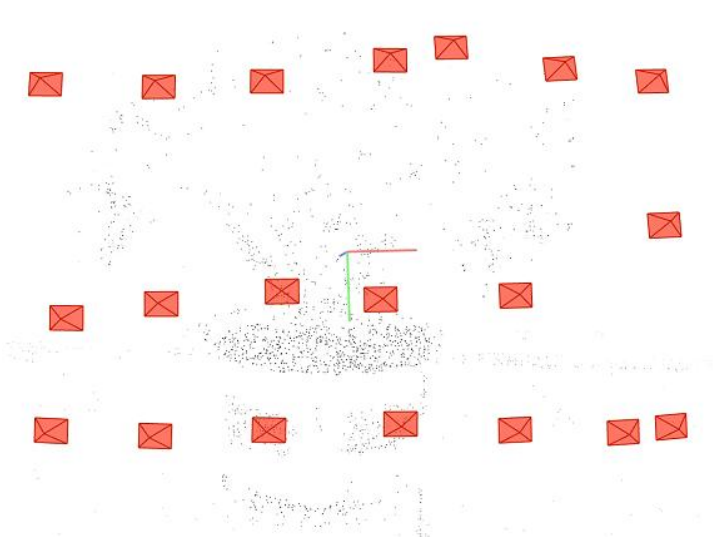}
         \caption{forward-facing scene}\vspace{4px}
         \label{fig:forward-facing}
     \end{subfigure}
     \hfill
     \begin{subfigure}[b]{0.5\linewidth}
         \centering
         \includegraphics[width=\textwidth]{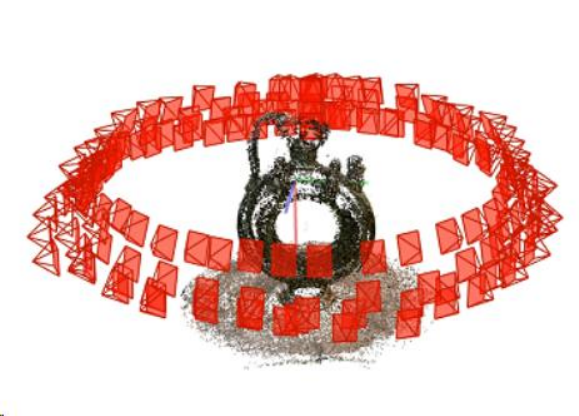}
         \caption{surrounding scene}\vspace{4px}
         \label{fig:surround}
     \end{subfigure}
        \caption{Shooting positions of forward-facing and surrounding scenes.}
        \label{fig:dataset_construction}
\end{figure}

\begin{figure*}[!htb]
\centering
\includegraphics[width=\linewidth]{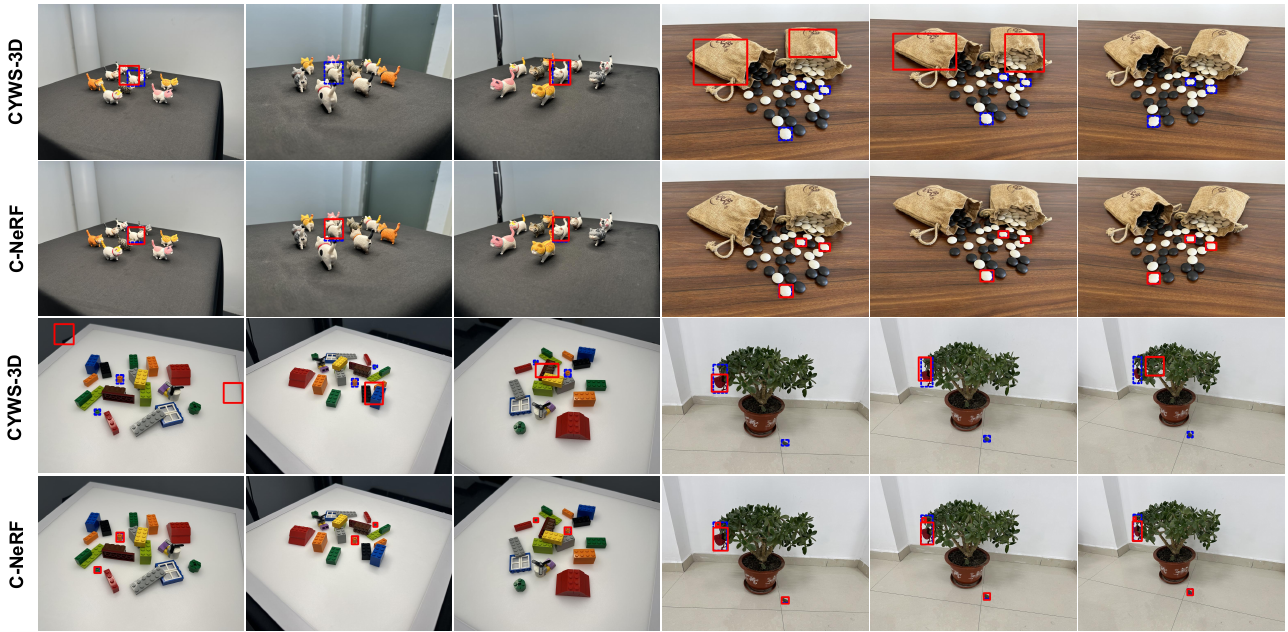}
\caption{Comparison of instance change detection method CYWS-3D~\cite{sachdeva2023change3D} with \textsc{C-NeRF} on different scenes. Red boxes are the predictions and blue boxes are the ground truth.}
\label{fig:object_level}
\end{figure*}

\begin{figure*}
  \centering
  \includegraphics[width=\linewidth]{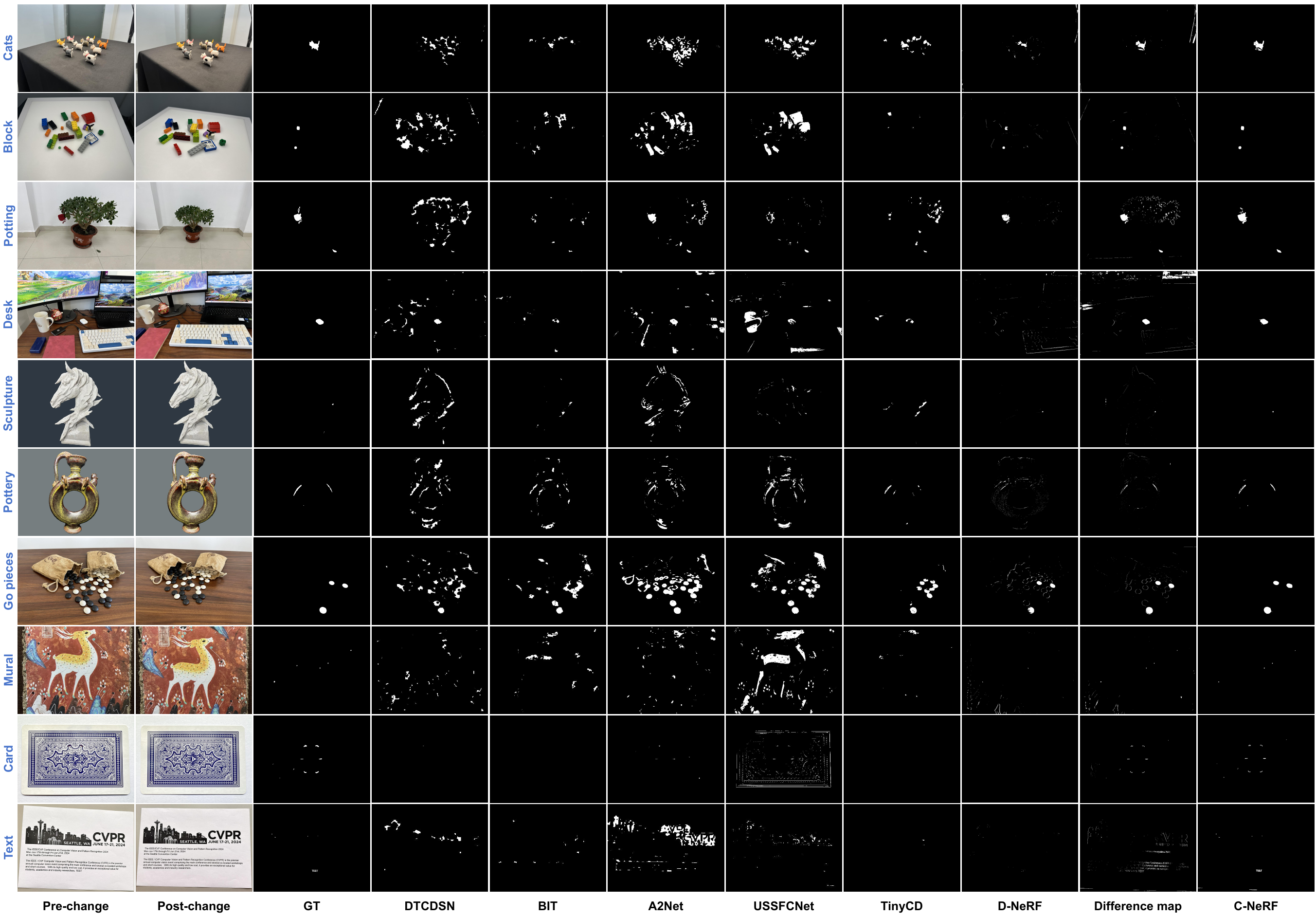}
  \caption{Some typical fine-grained change detection results of different methods on different scenes.}
  \label{fig:pixel_level}
\end{figure*}

\textbf{Evaluation metrics.}
We evaluate instance change detection using mean average precision (MAP) and pixel-level change detection using Precision (P), Recall (R), F1-measure (F1), and Intersection over Union (IoU).

\subsection{Results}

\textbf{Instance change detection.}
As CYWS-3D \cite{sachdeva2023change3D}  cannot detect fine-grained changes, we conducted the instance change detection experiments on instance change scenes, including Cats, Blocks, Potting and Go Pieces. 
To generate our instance change detection results, we use the tightest bounding box of the change region detected by \textsc{C-NeRF}. 
%


As shown in Fig.~\ref{fig:object_level}, CYWS-3D can detect large change objects such as the white cat and the bag hanging on the tree but fails to detect small objects such as go pieces and blocks. In contrast, our method, \textsc{C-NeRF}, can accurately detect all changes in these four scenes.

In Table~\ref{table:res_object-level}, we can see that CYWS-3D achieves a MAP value of 15.9 on the Cats scene and 1.5 on the Go pieces scene, which is not as high as the MAP values achieved by \textsc{C-NeRF}. For instance, \textsc{C-NeRF} achieves much higher MAP values of 85.1 and 87.3 on these respective scenes. Moreover, CYWS-3D attains 0 MAP values on the Blocks and Potting scenes, whereas \textsc{C-NeRF} achieves 67.9 and 34.5 MAP values on these two scenes, respectively. These quantitative results clearly demonstrate the superiority of the \textsc{C-NeRF} model over CYWS-3D.


\begin{table}[!t]
\centering
\setlength{\belowcaptionskip}{0.0cm}
\caption{Quantitative comparison of CYWS-3D~\cite{sachdeva2023change3D} with \textsc{C-NeRF}. The best MAP values are highlighted in \textbf{bolded}.}
\resizebox{0.8\linewidth}{!}{
\begin{tabular}{c|cccc}
\toprule
Scene  & Cats & Blocks & Potting & Go pieces \\ \midrule
CYWS-3D \cite{sachdeva2023change3D}& 15.9 & 0 & 0 & 1.5   \\
\textsc{C-NeRF}   & \textbf{85.1} & \textbf{67.9} & \textbf{34.5}    & \textbf{87.3} \\
\bottomrule
\end{tabular}
}\vspace{-10px}
\label{table:res_object-level}
\end{table}

\textbf{Fine-grained change detection.}
For fine-grained change detection, we choose 2D change detectors DTCDSN~\cite{liu2020building}, BIT~\cite{chen2021remote}, A2Net~\cite{li2023lightweight} and USSFC-Net~\cite{lei2023ultralightweight}, and TinyCD~\cite{codegoni2023tinycd}. We train all the change detectors with $1, 200$ training image pairs. Besides, we modify D-NeRF~\cite{pumarola2021d} to generate aligned image pairs to calculate the change maps. We train D-NeRF with pre-change and post-change images as two time periods. Then we generate aligned image pairs at the given view directions of two time points. The final change maps of D-NeRF are generated by the absolute difference of the image pair with the best thresholds. We also generate the difference map with the aligned image pairs by \textsc{C-NeRF}.



%
As shown in Fig.~\ref{fig:pixel_level},  we can find that almost all the compared methods produce fake changes on the object boundaries due to misaligned images. DTCDSCN, BIT, A2Net, and TinyCD failed to detect tiny changes of the Mural (see the 8th row of Fig.~\ref{fig:pixel_level}) and Card (see the 9th row of Fig.~\ref{fig:pixel_level}). A2Net can identify large changes but with low precision (see the 2nd, 3rd and 4th rows of Fig.~\ref{fig:pixel_level}). TinyCD can identify changes, but they are not complete (see the 1st and 6th rows of Fig.~\ref{fig:pixel_level}). 
D-NeRF fails to detect the changes of the planner scenes in the 8th, 9th and 10th rows of Fig.~\ref{fig:pixel_level}.
In contrast, \textsc{C-NeRF} can identify real changes on different scenes while filtering out the fake ones.


As shown in Table~\ref{table:res_fine_grained}, TinyCD achieves the best performances among the compared methods on NeRFCD. While more complex models, BIT and DTCDSN, show worse performance than TinyCD as they need more training data. 
Although D-NeRF can align pre- and post-change images, it obtains higher precision and lower recall than TinyCD. 
With the aligned image pairs, the simple absolute difference-based method achieves 46.52 and 33.81 F1 and IoU values, which are higher than other compared methods. 
The F1 and IoU values of \textsc{C-NeRF} are 74.48 and 62.18, which rank at the first place. The quantitative results demonstrate that \textsc{C-NeRF}  is capable of capturing fine-grained changes in various scenes.


\begin{table}[]
\centering
\caption{Quantitative comparison of 2D change detection with \textsc{C-NeRF}. The best values are highlighted in \textbf{bolded}.}
\resizebox{0.8\linewidth}{!}{
\begin{tabular}{l|cccc}
\toprule
Method  & P$\uparrow$ & R$\uparrow$ & F1 $\uparrow$ & IoU $\uparrow$ \\ 
\midrule
DTCDSN~\cite{liu2020building} & 8.47 & 53.34 & 14.61 & 7.88   \\
BIT~\cite{chen2021remote} & 20.97 & 48.81& 29.34 & 17.19   \\
A2Net~\cite{li2023lightweight}  & 11.41 & 78.29 & 19.92 & 11.06   \\
USSFC-Net~\cite{lei2023ultralightweight} & 10.14 & 72.21 & 17.85 & 9.8   \\
TinyCD~\cite{codegoni2023tinycd} & 34.49 & 47.62 & 40.02 & 25.01   \\
D-NeRF~\cite{pumarola2021d} & 35.68 & 41.92 & 35.22 & 24.32   \\
\hline
Difference & 42.61 & 67.46 & 46.52 & 33.81   \\
\textsc{C-NeRF}    & \textbf{81.2} & \textbf{75.1} & \textbf{74.48}    & \textbf{62.18} \\
\bottomrule
\end{tabular}
}\vspace{-10px}
\label{table:res_fine_grained}
\end{table}


\subsection{Discussion}

\begin{figure}[!t]
  \centering
  \setlength{\belowcaptionskip}{-0.4cm}
  \includegraphics[width=\linewidth]{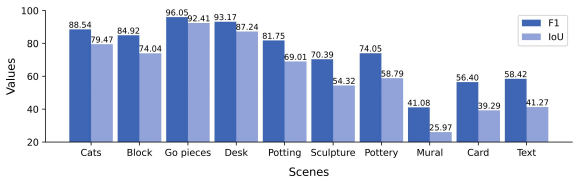}
  \caption{Performance of \textsc{C-NeRF} on different scenes.}  \label{fig:res_of_different_scenes}
\end{figure}

\textbf{Performance of \textsc{C-NeRF} on different scenes.}
Fig.~\ref{fig:res_of_different_scenes} shows the F1 and IoU values of \textsc{C-NeRF} on different scenes. For general scenes with large changes, such as Cats, Desk, and Go pieces, \textsc{C-NeRF} can achieve higher performances. While for the scenes with tiny changes, such as Mural, Card, and Text, \textsc{C-NeRF} obtains lower F1 and IoU values. This might be caused by the rendering quality of NeRF. Low rendering quality blurs the boundaries of the change regions, which reduces the areas of the changes. Thus, the change regions of \textsc{C-NeRF} are small of the regions of ground truth.

\textbf{Running time.} We use 186 image pairs to train the NeRF models and render 120 change maps. As shown in able~\ref{table:running_time}, aligning NeRFs is the most time-consuming process.


\begin{table}[!h]
\centering
\caption{Time consumption of main stages of \textsc{C-NeRF}.}
\resizebox{\linewidth}{!}{
\begin{tabular}{c|cccc}
\toprule
Stage  & NeRF alignment & Change point detection & Rendering \\ \midrule

Time (Hour) & 16.25 & 0.75 & 2.6 \\
\bottomrule
\end{tabular}
}
\vspace{-10px}
\label{table:running_time}
\end{table}



\textbf{Pros and cons of \textsc{C-NeRF}.} There are three main advantages of \textsc{C-NeRF}: \ding{182} \textsc{C-NeRF} eliminates the need for high-precision camera pose and image alignment, making scene shooting more accessible for ordinary people.
\ding{183} \textsc{C-NeRF} breaks the traditional pipeline of 2D change detection, introduces a novel change detection framework.
\ding{184} With \textsc{C-NeRF}, we are able to synthesize change maps under arbitrary specified view directions, unlike 2D change detections which only identify changes in one view direction.
However, there are some limitations: \ding{182} Large scenes like remote sensing and street view are not considered in this paper. Further study is needed on how to conduct change detection on these scenes. \ding{183} The performance of \textsc{C-NeRF} is heavily influenced by NeRF \cite{mildenhall2020nerf}. For example, \textsc{C-NeRF}
cannot deal with the changes in the specular regions and has a long training time. Using different NeRF models might extend the capabilities of \textsc{C-NeRF}, which will be studied in our future work.
\section{Conclusion}

In this work, we have identified a challenging but meaningful task, \ie, 3D change detection based on NeRF representation, which aims to synthesize change maps of a scene under arbitrary specified view directions. 
To this end, we studied the naive solution based on existing NeRF methods and found that this method is not able to identify the change regions accurately.
To address the limitations, we proposed a novel method denoted as \textsc{C-NeRF} that contains three modules to address the challenges of the naive solutions.
Specifically, we first performed spatial alignment to build two NeRFs for two image sets captured under two different timestamps. 
Then, we proposed to add a direction-consistent constraint to filter noise change points.
Finally, we detailed the rendering method to generate the change map under a view direction.
We build a new dataset, \textsc{NeRFCD}, with 10 different scenes.
Extensive experiment results have demonstrated that our method is effective in detecting instance-level and fine-grained changes in various scenes, outperforming a series of NeRF-based 2D change detection methods significantly.

{
    \small
    \bibliographystyle{ieee_fullname}
    \bibliography{egbib}
}
\vspace{10px}

In this supplementary material, we provide the code link of \href{ https://github.com/C-NeRF/C-NeRF}{\textsc{C-NeRF}}, dataset details, quantitative results of different change detectors on different scenes,  more visualization results of \textsc{C-NeRF}, and a video demo.

\section{Dataset Details}
We capture all scenes with iPhone 14 Pro and set the resolution of the images to $1008\times756$ to train D-NeRF and \textsc{C-NeRF}. To train 2D change detectors, we resize the images into $512\times 512$. The detailed number of images for each scene is shown in Table.~\ref{table:dataset_detail}.





\begin{table}[!h]
\centering
\caption{Dataset details.}
\resizebox{\linewidth}{!}{
\begin{tabular}{c|cccccc}
\toprule
Scene  & Pre-change & Post-change & Render & Training &Testing\\ 
\midrule
Cats  & 118 & 125 &  120 & 120 & 10  \\
Block & 186 & 187 &  120 & 120 & 10\\
Potting & 21 & 21  & 120 & 120 & 10   \\
Desk  & 20 & 20  & 120 & 120 & 10 \\
Sculpture  & 26 & 30 & 120 & 120 & 10 \\
Pottery  & 26 & 22 & 120 & 120 & 10 \\
Go pieces & 22 & 26 & 120 & 120 & 10   \\
Mural & 22 & 18 & 120 & 120 & 10   \\
Card & 27 & 27 &  120 & 120 & 10   \\
Text & 17 & 20 &  120 & 120 & 10  \\
\bottomrule
\end{tabular}
}
\label{table:dataset_detail}
\end{table}

\section{Quantitative results of different change detectors on different scenes}
We report a quantitative comparison of D-NeRF, results of difference maps with aligned image pairs generated by \textsc{C-NeRF}, and \textsc{C-NeRF} in different scenes in Table~\ref{table:results_detail}. We can find that \textsc{C-NeRF} achieves the highest F1 and IoU values in all ten scenes. This demonstrates that \textsc{C-NeRF} is more suitable for detecting changes in various scenes.



\section{More visualization results of \textsc{C-NeRF}}
\begin{figure*}[!ht]
    \centering
    \captionsetup{font={small}}
    \includegraphics[width=0.87\linewidth]{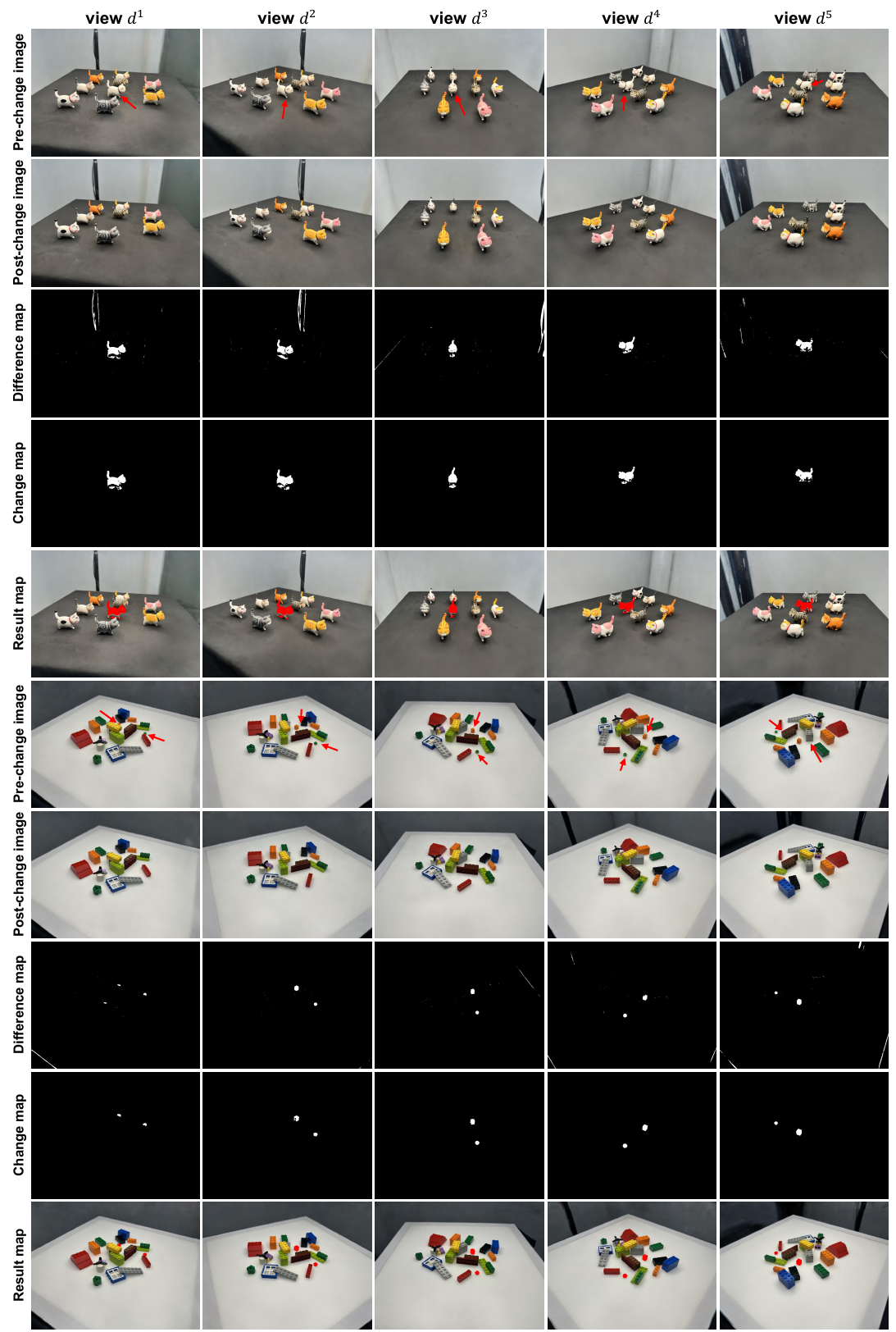}
    \caption{Change detection results of \textsc{C-NeRF} of different viewpoints of Cats and Blocks. The change objects are pointed by the red arrows.}
    \label{fig:results1}
\end{figure*}

\begin{figure*}[!ht]
    \centering
    \captionsetup{font={small}}
    \includegraphics[width=0.87\linewidth]{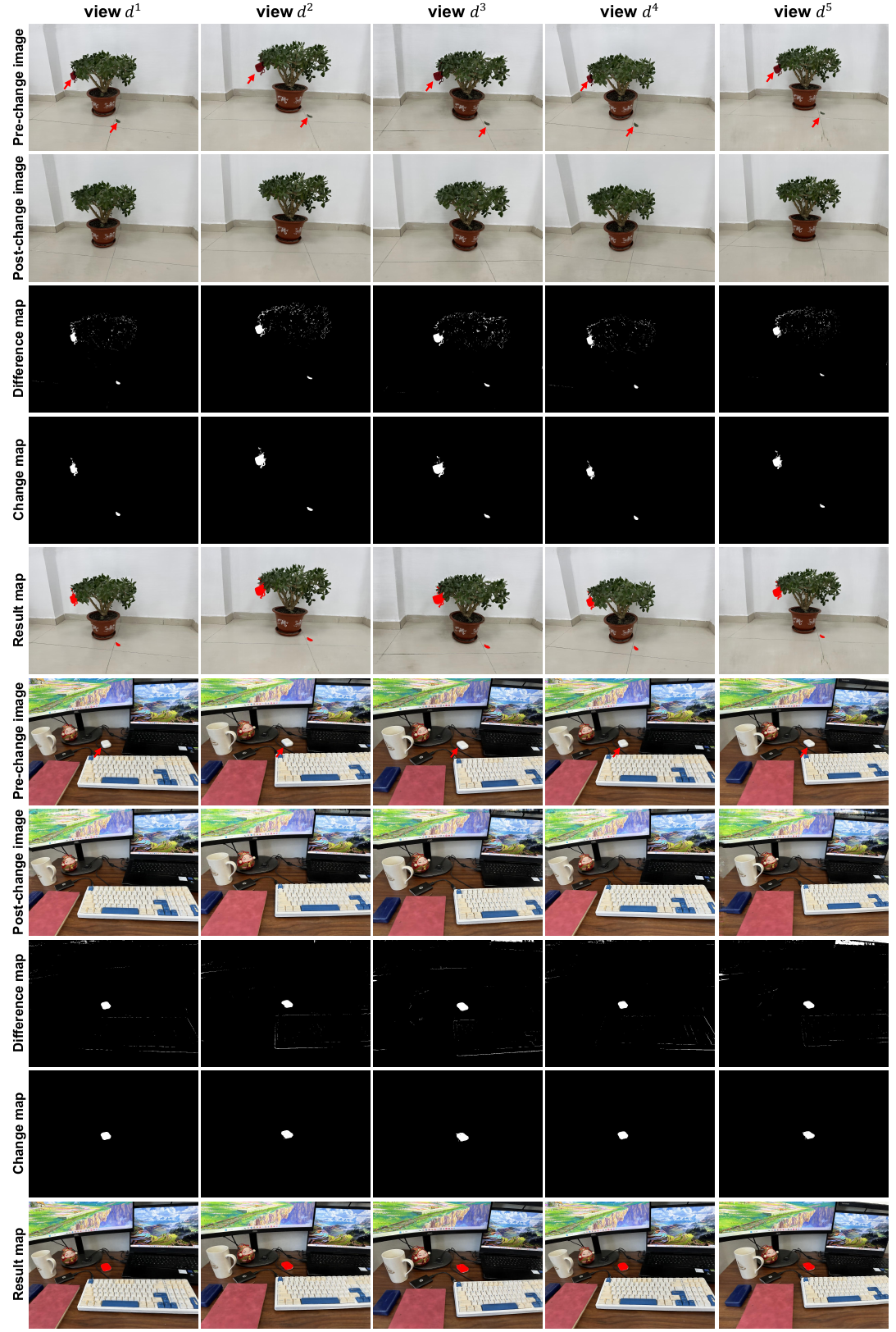}
    \caption{Change detection results of \textsc{C-NeRF} of different viewpoints of Potting and Desk. The change objects are pointed by the red arrows}
    \label{fig:results2}
\end{figure*}

\begin{figure*}[!ht]
    \centering
    \captionsetup{font={small}}
    \includegraphics[width=0.87\linewidth]{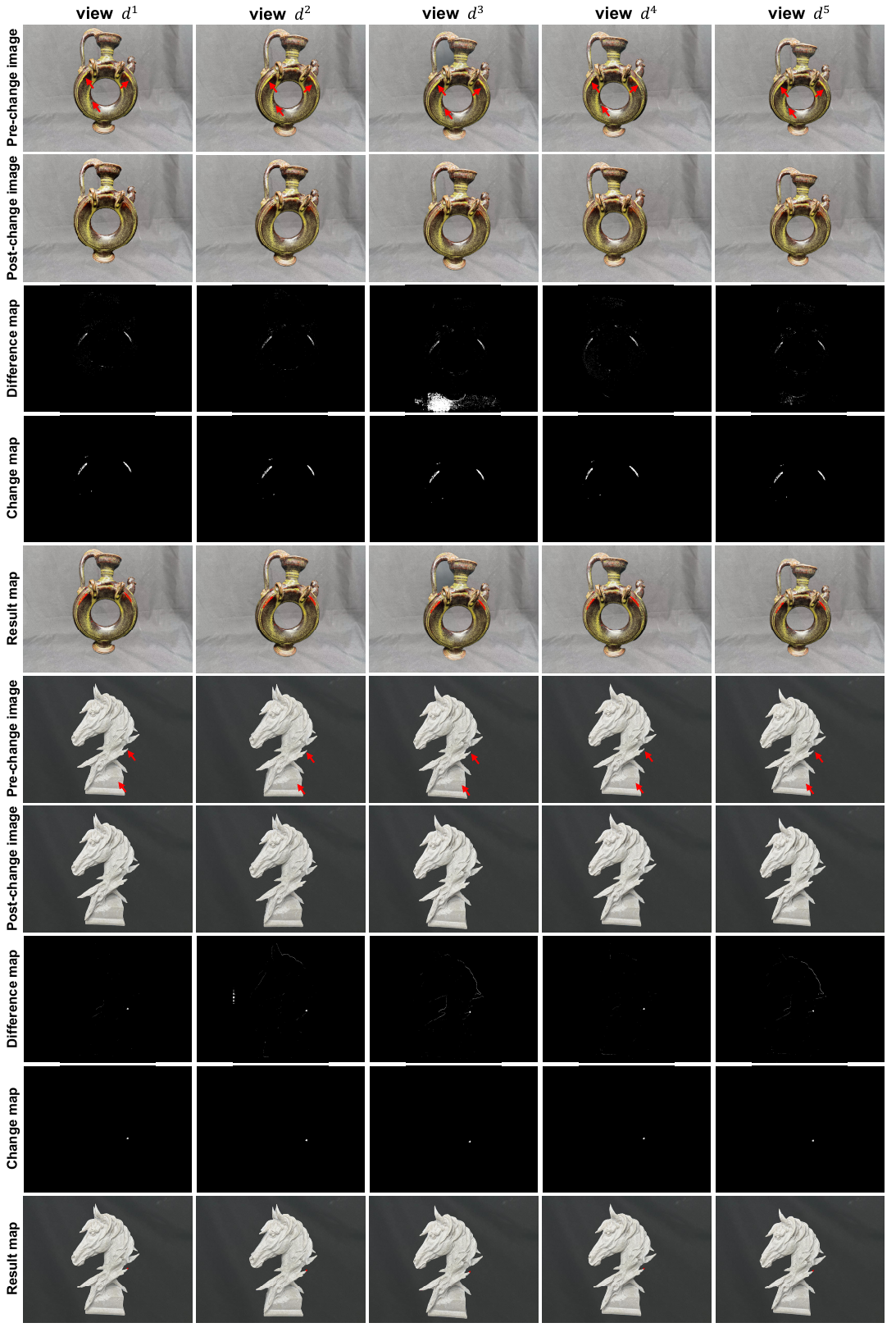}
    \caption{Change detection results of \textsc{C-NeRF} of different viewpoints of Pottery and Sculpture. The change objects are pointed by the red arrows}
    \label{fig:results3}
\end{figure*}

\begin{figure*}[!ht]
    \centering
    \captionsetup{font={small}}
    \includegraphics[width=0.87\linewidth]{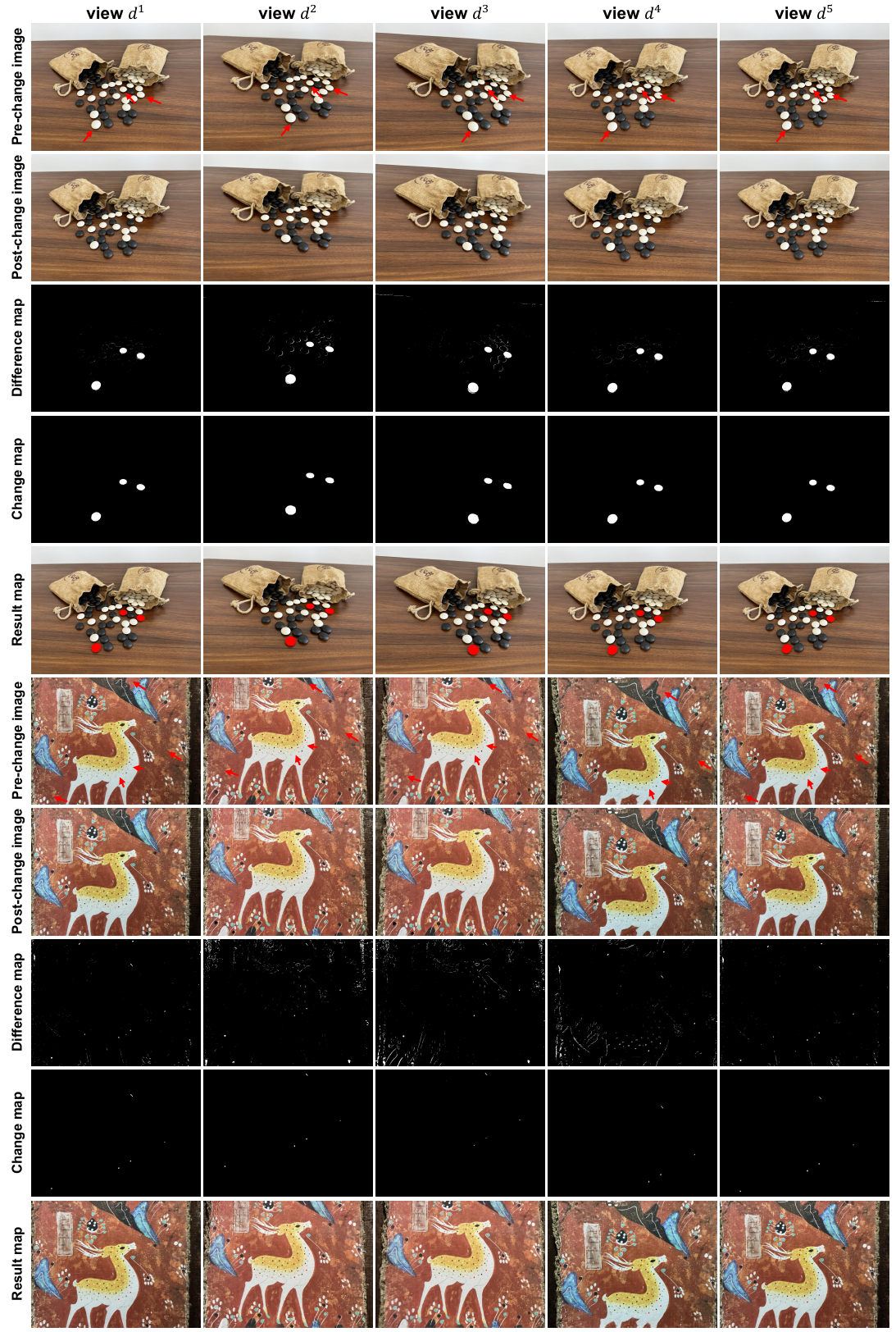}
    \caption{Change detection results of \textsc{C-NeRF} of different viewpoints of Go pieces and Mural. The change objects are pointed by the red arrows}
    \label{fig:results4}
\end{figure*}

\begin{figure*}[!ht]
    \centering
    \captionsetup{font={small}}
    \includegraphics[width=0.87\linewidth]{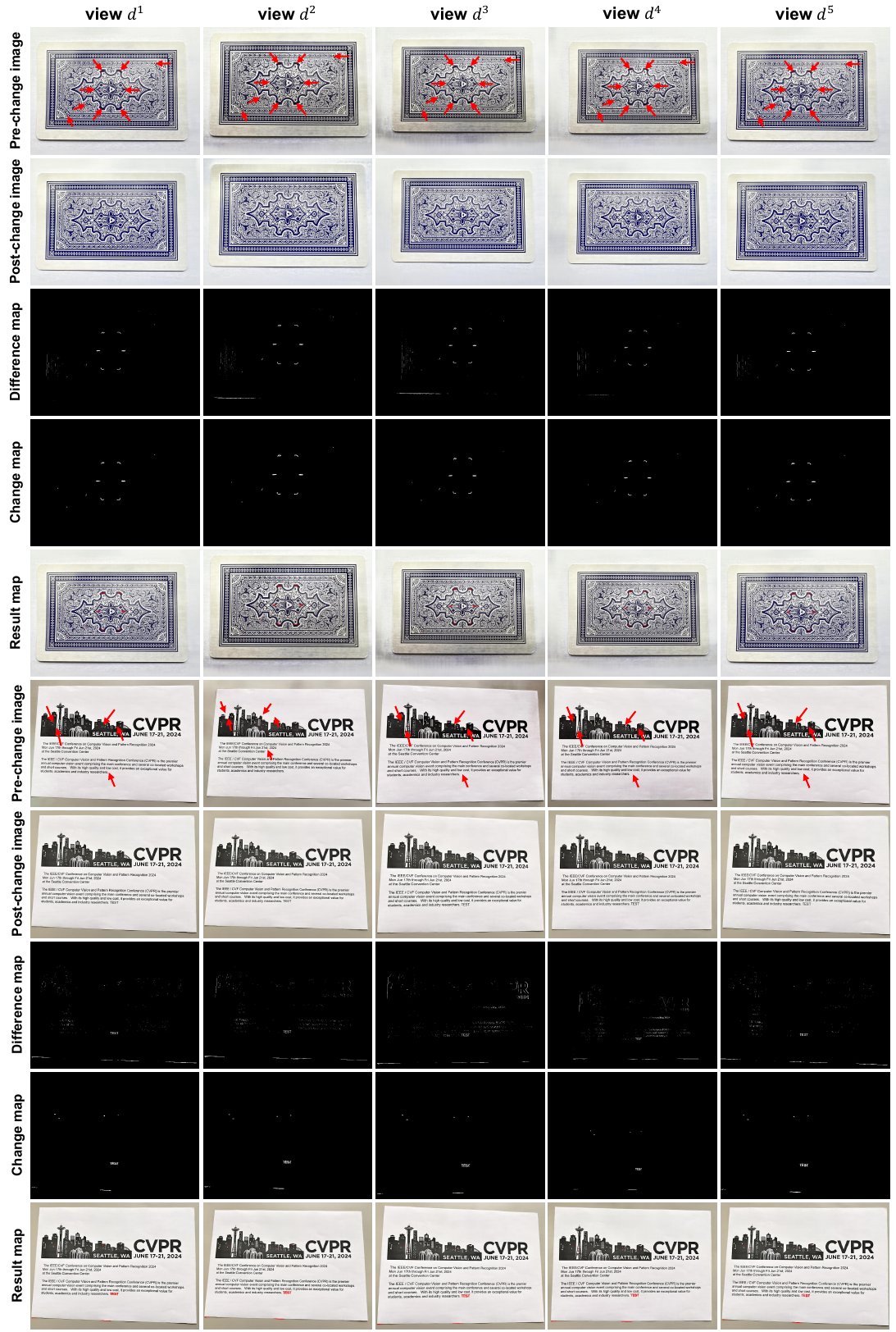}
    \caption{Change detection results of \textsc{C-NeRF} of different viewpoints of Card and Text. The change objects are pointed by the red arrows}
    \label{fig:results5}
\end{figure*}

To demonstrate the effectiveness of \textsc{C-NeRF} in detecting changes in different viewpoints, we provide more visualization results in Figs.~\ref{fig:results1}, \ref{fig:results2}, \ref{fig:results3}, \ref{fig:results4}, and \ref{fig:results5}. For each scene, we provide detection results of five view directions with intervals of ten. From the figures, we find that \textsc{C-NeRF} can detect the changes of each scene under different viewpoints. 

\section{Video demo}
In order to better showcase our results, we have provided images including pre-change image, post-change image, difference maps, and results of \textsc{C-NeRF} in each video frame synchronously. Please see the attached video.

\begin{table}[t]
\centering
\caption{Quantitative comparison of D-NeRF, results of difference maps with aligned image pairs generated by \textsc{C-NeRF}, and \textsc{C-NeRF} in different scenes. The best values are \textbf{bolded}.}
\resizebox{\linewidth}{!}{ 
\begin{tabular}{c|c|cccc}
\toprule
Scene & Method & P$\uparrow$ & R$\uparrow$ & F1$\uparrow$ & IoU$\uparrow$ \\ 
\midrule
\multirow{3}{*}{Cats} & D-NeRF & 59.24 & 60.57 & 59.68 & 42.71 \\
& Difference & 58.80 & 83.5 & 68.86 & 52.7 \\
\multirow{3}{*}{ } & \textsc{C-NeRF}  & \textbf{86.70}  & \textbf{90.46}& \textbf{88.54} & \textbf{79.47}\\
\cmidrule(r){1-6}
\multirow{3}{*}{Blocks} & D-NeRF & 25.83 & 79.48 & 38.06 & 23.88\\
& Difference  & 50.1 & 91.17 & 64.3 & 48.22  \\
\multirow{3}{*}{ } & \textsc{C-NeRF}  & \textbf{77.70}  & \textbf{93.62} & \textbf{84.92} & \textbf{74.04} \\
\cmidrule(r){1-6}
\multirow{3}{*}{Go pieces}  & D-NeRF & 46.34 & 80.27 & 58.41 & 41.42 \\
& Difference & \textbf{99.97} & 17.69 & 30.04 & 17.68 \\
\multirow{3}{*}{ } & \textsc{C-NeRF}  & 92.66 & \textbf{99.71} & \textbf{96.05} & \textbf{92.41} \\
\cmidrule(r){1-6}
\multirow{3}{*}{Desk} & D-NeRF  & 20.81 & 41.72 & 27.74 & 16.42 \\
& Difference  & 19.62 & 86.09 & 31.67 & 18.96 \\
\multirow{3}{*}{ } & \textsc{C-NeRF}  & \textbf{96.02} & \textbf{90.48} & \textbf{93.17} & \textbf{87.24} \\
\cmidrule(r){1-6}
\multirow{3}{*}{Potting} & D-NeRF  & \textbf{83.87} & 70.77 & 76.71 & 62.23 \\
& Difference  & 44.97 & 89.42 & 59.81 & 42.68 \\
\multirow{3}{*}{ } & \textsc{C-NeRF}  & 71.59 & \textbf{94.66} & \textbf{81.75} & \textbf{69.01} \\
\cmidrule(r){1-6}
\multirow{3}{*}{Sculpture} & D-NeRF  & 69.5 & 38.78 & 49.68 & 33.08 \\
& Difference  & 19.47 & 53.42 & 28.53 & 16.6 \\
\multirow{3}{*}{ } & \textsc{C-NeRF}  & \textbf{92.48} & \textbf{56.81} & \textbf{70.39} & \textbf{54.32} \\
\cmidrule(r){1-6}
\multirow{3}{*}{Pottery} & D-NeRF  & 7.51 & 9.3 & 8.25 & 4.3 \\
& Difference  & 57.99 & 37.29 & 45.36 & 29.33 \\
\multirow{3}{*}{ } & \textsc{C-NeRF}  & \textbf{90.08} & \textbf{62.86} & \textbf{74.05} & \textbf{58.79} \\
\cmidrule(r){1-6}
\multirow{3}{*}{Mural} & D-NeRF & 4.56 & 7.7 & 5.71 & 2.95 \\
& Difference & 5.92 & \textbf{29.32} & 9.74 & 5.15 \\
\multirow{3}{*}{ } & \textsc{C-NeRF}  & \textbf{84.08} & 27.2 & \textbf{41.08} & \textbf{25.97} \\
\cmidrule(r){1-6}
\multirow{3}{*}{Card} & D-NeRF  & 36.84 & 22.05 & 26.02 & 15.2 \\
& Difference  & \textbf{79.19} & 39.61 & 52.8 & 35.87 \\
\multirow{3}{*}{ } & \textsc{C-NeRF}  & 77.69 & \textbf{44.28} & \textbf{56.40} & \textbf{39.29}  \\
\cmidrule(r){1-6}
\multirow{3}{*}{Text} & D-NeRF  & 2.26 & 8.6 & 1.92 & 0.98 \\
& Difference  & 8.54 & 66.56 & 15.05 & 8.18 \\
& \textsc{C-NeRF}  & \textbf{43.04} & \textbf{90.92} & \textbf{58.42} & \textbf{41.27} \\
\bottomrule
\end{tabular}
}
\label{table:results_detail}
\end{table}

\end{document}